\documentclass[runningheads]{llncs}
\usepackage{graphicx}
\newcommand{\citeauthor}[1]{\cite{#1}}
\newcommand{\citeyear}[1]{\cite{#1}}

\usepackage{amsfonts}
\usepackage{ dsfont }
\usepackage{amsmath}
\usepackage{verbatim}
\DeclareMathOperator*{\argmax}{argmax}

\usepackage{algorithm2e}
\usepackage{booktabs}

\usepackage{tikz}
\usepackage{caption}
\usetikzlibrary{positioning}
\tikzset{%
  every neuron/.style={
    circle,
    draw,
    minimum size=1cm
  },
  neuron missing/.style={
    draw=none, 
    scale=2,
    text height=0.15cm,
    execute at begin node=\color{black}$\vdots$
  },
  neuron hmissing/.style={
    draw=none, 
    scale=3,
    text height=0.15cm,
    execute at begin node=\color{black}$\hdots$
  },
}

\usepackage[toc,page]{appendix}

\usepackage{todonotes}

\renewcommand{\figurename}{Figure}

\newcommand{\getmydate}{%
  \ifcase\month%
    \or Januar\or Februar\or M\"arz%
    \or April\or Mai\or Juni\or Juli%
    \or August\or September\or Oktober%
    \or November\or Dezember%
  \fi\ \number\year%
}

\newcounter{equationset}
\begin{document}
\title{Deep Ordinal Reinforcement Learning}
%\thanks{This work was funded by DFG under project number 520 01481.}}
\author{Alexander Zap \and
Tobias Joppen\and %\inst{1}
Johannes F\"urnkranz} %\inst{1}
\authorrunning{A. Zap et al.}
\institute{TU Darmstadt, 64289 Darmstadt, Germany\\
\email{alexander.zap@stud.tu-darmstadt.de}\\
\email{\{tjoppen, juffi\}@ke.tu-darmstadt.de}
}
\maketitle              % typeset the header of the contribution
\begin{abstract}
Reinforcement learning usually makes use of numerical rewards, which have nice properties but also come with drawbacks and difficulties.
Using rewards on an ordinal scale (ordinal rewards) 
%or preferences 
is an alternative to numerical rewards that has received more attention in recent years.
%, not only limited to RL.
In this paper, a general approach to adapting reinforcement learning problems to the use of ordinal rewards is presented and motivated.
We show how to convert common reinforcement learning algorithms to an ordinal variation by the example of Q-learning and introduce Ordinal Deep Q-Networks, which adapt deep reinforcement learning to ordinal rewards.
%by the example of Deep Q-Network.
Additionally, we run evaluations on problems provided by the OpenAI Gym framework, showing that our ordinal variants exhibit a performance that is comparable to
the numerical variations for a number of problems.
We also give first evidence that our ordinal variant is able to produce better results for problems with less engineered and simpler-to-design reward signals.

\end{abstract}
%
%   The maximum length of papers is 16 pages in this format!
%

    \section{Introduction}

        Conventional reinforcement learning (RL) algorithms rely on numerical feedback signals.
        %and the most prominent algorithms and applications use this kind of feedback signals.
		Their main advantages include the ease of aggregation, efficient gradient computation, and many use cases where numerical reward signals come naturally, 
		%an intuitive meaning where maximizing the numerical reward is natural, 
		often representing a quantitative property.
		However, in some domains numerical rewards are hard to define and are often subject to certain problems. 
		One issue of numerical feedback signals is the difficulty of \emph{reward shaping}, which is the task of creating a reward function. 
		Since RL algorithms use rewards as direct feedback to learn a behavior which optimizes the aggregation of received rewards, the reward function has a significant impact on the behavior that is learned by the algorithm. 
		Manual creation of the reward function is often expensive, non-intuitive and difficult in certain domains and can therefore cause a bias in the optimal behavior that is learned by an algorithm.
		Since the learned behavior is sensitive to these reward values, the rewards of an environment should not be introduced or shaped arbitrarily if they are not explicitly known or naturally defined.
		This can also lead to another problem called \emph{reward hacking}, where algorithms are able to exploit a reward function and miss the intended goal of the environment, caused by being able to receive better rewards through undesired behavior. 
		The use of numerical rewards furthermore requires \emph{infinite rewards} to model undesired decisions in order to not allow trade-offs for a given state. 
		This can be illustrated by an example in the medical domain, where it is undesirable to be able to compensate one occurrence of \emph{death of patient} with multiple occurrences of \emph{cured patient} to stay at a positive reward in average and therefore artificial feedback signals are used that can not be averaged.
%	
%		\subsection{Ordinal rewards for Reinforcement Learning}
		These issues 
		%of defining and using numerical rewards that were mentioned above
		have motivated the search for alternatives, such as preference-based feedback signals \cite{Wirth2017}.

        In this paper, we investigate the use of rewards on an ordinal scale, where we have information about the relative order of various rewards, but not about the magnitude of the quality differences between different rewards.
		Our goal is to extend reinforcement learning algorithms so that they can make use of ordinal rewards as an alternative feedback signal type 
		% for reinforcement learning 
		in order to avoid and overcome the problems with numerical rewards.
		
		Reinforcement learning with ordinal rewards has multiple advantages and directly addresses multiple issues of numerical rewards.
		Firstly, the problem of \emph{reward shaping} is minimized, since the manual creation of the ordinal reward function specifically by the reward ordering often is intuitive and can be done easily without the need of exact specifications for reward values.
		Even though the creation of ordinal reward values, so-called reward tiers, through the ascending order of feedback signals introduces a naturally defined bias, it omits the largely introduced artificial bias by the manual shaping of reward values.
		At the same time, ordinal rewards %thereby simultaneously
		simplify the problem of \emph{reward hacking} because the omission of specific numeric reward values has the effect that any possible exploitation of rewards by an algorithm is only dependent on an incorrect reward order, which can be more easily fixed than the search for correct numerical values.
		While the use of \emph{infinite rewards} can not be modelled directly, it is still possible to define infinite rewards as highest or lowest ordinal reward tier, and implement policies which completely avoid and encourage certain tiers.
		
		Since the creation of the ordinal reward function is cheap and intuitive, it is especially suitable for newly defined environments since it enables the easy definition of ordinal rewards by ordering the possible outcomes naturally by desirability.
		Additionally it should be noted that for existing environments with numerical rewards it is possible to extract ordinal rewards from these environments.
		
		The focus of this paper is the technique of using ordinal rewards for reinforcement learning.
		To this end, we propose an alternative reward aggregation for ordinal rewards, introduce a method for policy determination from ordinal rewards and 
	    compare the performance of ordinal reward algorithms to algorithms for numerical rewards.
		In Section~\ref{related work}, we discuss related work and previous approaches. % to this topic.
		A formal definition of common reinforcement learning terminology can be found in Section~\ref{definition}.
		Section~\ref{ordinal definition} introduce reinforcement learning algorithms which use ordinal reward aggregations instead of numerical rewards, and illustrates the differences to conventional approaches.  
		In Section~\ref{experiments} experiments are executed on the framework of OpenAI Gym and common reinforcement learning algorithms are compared to ordinal reinforcement learning.

% \todo[inline]{Give a short overview of the organization of this paper (approx.\ 1 sentence per section. Maybe before that briefly summarize the main contribution of the paper. Ideally, this fills this page...}

    \section{Related Work}
    \label{related work}
	The technique of using rewards on an ordinal scale as an alternative to numerical rewards is mainly based on the approach of preference learning (PL) \cite{plbook}. 
	% Common machine learning techniques define a function $f(x) = y$, which is learned by observing multiple labeled data instances for training, that afterwards is able to predict numerical or nominal labels $y$ with the feature input $x$ of unlabeled data.
	In contrast to traditional supervised learning, PL follows the core idea of having preferences over states or symbols as labels and predicting these preferences as the output on unseen data instances instead of labelling data with explicit nominal or numerical values.
	% This means that the function defined by the preference learning technique uses labels $y \in [y_1, ..., y_n]$ with a preference definition of $y_i \succ y_j \text{ or } y_j \succ y_i \text { for every } y_i, y_j \in y, i \neq j$.
	% Following approaches to use preference-based reward types as the feedback signal in reinforcement learning have been made in related work before.

    Recently, there have been several proposals for combining PL with RL,
	%The approach of directly combining PL and RL, also called \emph{preference-based reinforcement learning}, has been proposed first by \citeauthor{Fuernkranz2012} and has been investigated extensively in \citeauthor{Wirth2017},
	where pairwise preferences over trajectories, states or actions are defined and applied as feedback signals in reinforcement learning algorithms instead of the commonly used numerical rewards.
	For a survey of such preference-based reinforcement learning algorithms, we refer the reader to \cite{Wirth2017}.
	
	While preference-based RL provides algorithms for learning an agent's behavior from pairwise comparison of trajectories, \citeauthor{Weng2013} presents an approach for creating preferences over multiple trajectories in the order of ascending ordinal reward tiers, thereby deviating from the concept of pairwise comparisons over trajectories.
	Using a tutor as an oracle, this approach approximates a latent numerical reward score from a sequence of received ordinal feedback signals.
	This alternative reward computation functions as a reward transformation from the ordinal to the numerical scale and is applicable on top of an existing reinforcement learning algorithm.
	% Unfortunately, this approach of ordinal reward transformation by a querying system requires explicit human feedback. 
	% With the work of \citeauthor{Daniel2016}, which proposes to learn an intermediate reward signal from expert ratings and use the learned signal to identify preferences over demonstrations, the number of required human input can be reduced but not completely eliminated.
	%
	% Therefore this thesis aims to implement a reinforcement learning algorithm for environments with feedback signals on an ordinal scale with multiple reward tiers without the need of manual feedback.
	
    Contrary to this approach, we do not use a tutor for the comparison of trajectories but can directly use ordinal rewards as a feedback signal.
    % for reinforcement learning algorithms because of its property of providing an alternative possibility to omit the natural metric of numerical rewards in reinforcement learning.
	% Additionally, while previous work produces preference information for RL algorithms from manual definition, we introduce an approach to automatically extract rewards on an ordinal scale from existing environments with numerical rewards using the approach of \citeauthor{Joppen2019}.
	% To transfer existing numerical feedback in environments into preference-based information we follow the approach of \citeauthor{Joppen2019}.
	% In comparison to other approaches this generates preference information directly from numerical feedback in existing environments.
	In order to use environments where numerical feedback already exists without the need for acquiring human feedback about the underlying preferences, we automatically extract rewards on an ordinal scale from existing environments with numerical rewards. 
	To this end, we adapt an approach that has been proposed for Monte-Carlo Tree Search \citeauthor{Joppen2019} to reinforcement learning.
	
	Furthermore, we handle ordinal rewards in a similar manner as previous approaches by directly using aggregated received ordinal rewards for comparing different options.
    The idea of direct comparison of ordinal rewards builds on the  works of \cite{Weng2011}, \cite{Weng2012}, \citeauthor{Weng2016} and \citeauthor{Joppen2019}, which provide criteria for the direct comparison of ordinal reward aggregations. 
	% By adapting the approach presented by \citeauthor{Joppen2019} of transferring the numerical reward maximization problem into a best-choice maximization problem to common RL algorithms, a policy can be learned that maximizes the chance of playing a dominating trajectory.
	We utilize the approach of \citeauthor{Joppen2019}, which transfers the numerical reward maximization problem into a best-choice maximization problem for an alternative computation of the value function for reinforcement learning from ordinal feedback signals.
	% Furthermore a method to aggregate received ordinal rewards iteratively over time is suggested.
%
%	Our approach is heavily inspired on 
	\citeauthor{Joppen2019} used this idea 
%	, where a similar concept has successfully been tested 
	for adapting Monte Carlo Tree Search to the use of ordinal rewards. 
	
%	Therefore one can see this paper as a translation of this concept into reinforcement learning.
%	In fact we principally do not try to advertise a fundamentally new RL algorithm with this paper but try to show the possibility and potential benefits of transferring rewards into an ordinal scale and adapting a number of existing reinforcement learning algorithms to ordinal rewards.
    In summary, we automatically transfer numerical feedback into preference-based feedback and propose a new conceptual idea to utilize ordinal rewards for reinforcement learning, which should not be seen as an alternative for the existing algorithms stated above.
	Hence, we do not compare the performance of our new approach to any of the algorithms that use additional human feedback, but to common RL techniques that use numerical feedback.

    \section{Markov Decision Process and Reinforcement Learning}
    \label{definition}
	In this section, we briefly recapitulate Markov decision processes and reinforcement learning algorithms.
%	is presented and it is elaborated, how Reinforcement Learning algorithms iteratively learn to solve environments that can be modelled by Markov Decision Processes.
%	The following definitions are mainly regarded as state-of-the-art concepts for RL and can be found in 
	Our notation and terminology is based on 
	\citeauthor{Sutton2018}.
		
	    \subsection{Value function and policy for Markov Decision Process}
		A \emph{Markov Decision Process} (MDP) is defined as a tuple of ($S$, $A$, $P$, $R$) with $S$ being a finite set of states, $A$ being a finite set of actions, $T$ being the transition function $S \times A \times S\rightarrow \mathbb{R}$ that models the probability of reaching a state $s'$ when action $a$ is performed in state $s$, and $R$ being the reward function $S \times A \times S \rightarrow \mathbb{R}$ which maps a reward $r$ from a subset of possible rewards $r \in \{r_1, ..., r_n\} \subset \mathbb{R}$ to executing action $a$ in state $s$ and reaching $s'$ in the process.
		For further work we assume that $T$ is deterministic and a transition always has the probability of 0 or 1.
        Furthermore it is assumed that each action $a \in A$ is executable in any state $s \in S$, hence the transition function is defined for every element in $S \times A \times S$.
%		
		% \subsection{Policy}
		% \label{policy}
		% \hfill \break
		A policy $\pi$ is the specification which decision to take based on the environmental state. In a deterministic setting, it is modeled as a mapping $\pi : S \rightarrow A$ which directly maps an environmental state $s$ to the decision $a$ which should be taken in this state.
		% In many approaches the case of of non-probabilistic policies $\pi : S \rightarrow A$ is used, which can directly map an environmental state to the decision which should be taken.
		% The quality of a policy can be rated by aggregating numerical feedback signals through a value function.
		% The optimal policy $\pi^*$ in a state $s$ is to choose the action $a$ that maximizes the value function $V_\pi(s)$.
%		
		% \subsection{Value function}
		% \label{value function}
%		\hfill \break
		The value function $V_\pi(s)$ represents the expected quality of a policy $\pi$ in state $s$ with respect to the rewards that will be received in the future.
		Value functions for numerical rewards are computed by the expectation of the discounted sum of rewards $\mathds{E}[R]$. 
		The value function $V_\pi(s)$ of a policy $\pi$ in an environmental state $s$ therefore can be computed by
		
		\begin{equation}
        V_\pi(s) = \mathds{E}[R], \;\;\; R = \sum_{t=0}^\infty{\gamma^t}{r_{t}}
		\label{eq:V-pi-star}\end{equation}
		
		\noindent where $R$ is the discounted sum of rewards when following policy $\pi$, $\gamma$ a discount factor, and $r_t$ the direct reward at time step $t$. 
		% Therefore different environmental states can be directly compared and rated by the comparison of their value functions $V_\pi$.
		The optimal policy $\pi^*$ in a state $s$ is the policy with the largest $V_\pi(s)$, which complies with the goal of an RL algorithm to maximize expected future reward.

	    \subsection{Reinforcement Learning}
	    \label{reinforcement learning}
		Reinforcement learning can be described as the task of learning a policy that maximizes the expected future numerical reward.
		% This agent performs in an environment that takes actions from the agent as an input and outputs a newly reached environmental state and immediate feedback signal for the previously executed action.
		The agent learns iteratively by updating its current policy $\pi$ after every action and the corresponding received reward from the environment.
		Furthermore, the agent may perform multiple training sessions, so-called episodes, in the environment.
		% where the environment is restarted after every time the agent reaches a terminal state.
		Using the previously defined formalism, this can be expressed as approximating the optimal policy iteratively with a function $\hat{\pi}$, by repeatedly choosing actions that lead to states $s$ with the highest estimated value function $V_{\hat{\pi}}(s)$.
		In the following section two common reinforcement learning algorithms are introduced.
		
			\subsubsection{Q-learning.}
			\label{qlearning}
			The key idea of the Q-learning algorithm \cite{Q-learning} is to estimate Q-values $Q(s, a)$, which estimate the expected future sum of rewards $\mathds{E}[R]$ when choosing an action $a$ in a state $s$ and following the optimal policy $\pi^*$ afterwards.
			% Therefore Q-values are similar to the value function $V_\pi(s)$ of policy $\pi^*$, but compute the value function not only with respect to the state $s$ but specifically consider the value of taking action $a$ in state $s$.
			Hence the Q-value can be seen as a measure of \emph{goodness} for a state-action pair $(s, a)$, and therefore, in a given state $s$, the optimal policy $\pi^*$ should select the action $a$ that maximizes this value in comparison to other available actions in that state.
			The approximated Q-values are stored and iteratively updated in a Q-table.
			The Q-table is updated after an action $a$ has been performed in a state $s$ and the reward $r$ and the newly reached state $s'$ is observed.
			The computation of the expected Q-value is done by
			
			% The Bellman equation defines the value function as
			
			% $$V(s) = \max_a (r(s, a) + \gamma V(s')$$
			 
			% This can be translated to Q-values by moving the $\max$-Operator inside the bracket (since the reward $r$ is determined for action $a$) and choosing the best possible Q-value in the new state $s'$ by taking the action $a'$ the maximizes the $Q(s', a')$, which can be modeled by the formula of
			
			\begin{equation}
			  \hat{Q}(s, a) = r(s, a) + \gamma  \max_{a'} Q(s', a')
			  \label{eq:Bellman}
			 	\end{equation}
			
			\noindent Following this so-called Bellman equation, every previously estimated Q-value is updated with the newly computed expected Q-value with the formula
			%Hence the formula for updating Q-values is
			
			\begin{equation}Q(s, a) = Q(s, a) + \alpha [ r(s, a) + \gamma \max_{a'} Q(s', a') - Q(s, a) ]
			\label{eq:q-update}
			\end{equation}
			
			\noindent where $\alpha$ represents a learning rate and $\gamma$ the discount factor.

			\subsubsection{Deep Q-Network.}
			\label{dqn}
			The original Q-learning algorithm is limited to very simple problems, because of the explicitly stored Q-table, which essentially memorizes the quality of each possible state-action pair independently. Thus it requires, e.g., that each state-action pair has to be visited a certain number of times in order to make a reasonable prediction for this pair.
		A natural extension of this method is to replace the Q-table with a learned Q-function, which is able to predict a quality value for a given, possibly previously unseen state-action pair. 
			The key idea behind the Deep Q-Network (DQN)
%			is based on Q-learning and was presented by
\citeauthor{Mnih2013,RL-Atari}
is to learn a
%			The main difference between Q-learning and DQN consists of the method of Q-value computation.
%			While Q-learning uses a Q-table to predict the Q-value $Q(s, a)$ for state $s$ and action $a$, the Q-values for DQNs are predicted by a continuous function which returns a vector of Q-values for all possible actions.
%			This 
			continuous function $Q^{DQN}(s)$ 
			in the form of 
%			is realized by 
			a deep neural network with $m$ input nodes, which represent the feature vector of $s$, and $n$ output nodes, each containing the Q-value of one action $a$.

			Neural networks can be iteratively updated to fit the output nodes to the desired Q-values.		
			The expected Q-value for a state-action pair is calculated in the same manner as defined 
			in \eqref{eq:Bellman} 
			with the difference that the Q-values are now predicted by the DQN, with one output node $Q^{DQN}_{a}(s)$ for each possible action $a$.
			Therefore \eqref{eq:Bellman} becomes %for updating the neural network is
			\begin{equation}
            \hat{Q}^{DQN}_{a}(s) = r(s, a) + \gamma \max_{a'} Q^{DQN}_{a'}(s')
			\label{eq:dqn-update}
			\end{equation}
			
			\noindent where $Q^{DQN}_{a}(s)$ represents the Q-value node of action $a$ in state $s$.
			% While Q-learning executes the process of updating $Q(s, a)$ towards a target Q-value by overwriting the respective entry in a finite Q-table, DQN updates $Q^{DQN}_a(s)$ by fitting the neural network to the target Q-value.
			% This fitting process is described in \citeauthor{Werbos1990} and uses the technique of \emph{backpropagation}.\\
			
			In order to optimize the learning procedure, DQN makes use of several optimizations such as  \emph{experience replay}, the use of a separate \emph{target and evaluation network}, and \emph{Double Deep Q-Network}.  More details on these techniques can be found in the following paragraphs.
			
			% These should be explained shortly in the following paragraphs.
				\paragraph{Experience replay.}
				\noindent Using a neural network to fit the Q-value of the previously executed state-action pair as described in \eqref{eq:dqn-update} leads to overfitting to recent experiences because of the high correlation between environmental states across multiple successive time steps, and the property of neural networks to overfit recently seen training data.
				% As a solution to this problem we use \emph{experience replay}, which was first presented by \citeauthor{Lin1992}.
				Instead of only using the previous state-action pair for fitting the DQN, experience replay \citeauthor{Lin1992} uses a memory $M$ to store previous experience instances $(s, a, r, s')$ and iteratively reuses a random sample of these experiences to update the network prediction at every time step.
				% The information that is stored in $M$ includes state $s$, chosen action $a$, received reward $r$ and the newly reached state $s'$.
				% This way instead of only using the previous state-action pair for fitting the DQN, multiple memories from $M$ are sampled randomly and used for learning instead.
				
				\paragraph{Target and evaluation networks.}
				\noindent Frequently updating the neural network, which is simultaneously used for the prediction of the expected Q-value, leads to unstable fitting of the network.
				Therefore these two tasks, firstly the prediction of the target Q-value for network fitting and secondly the prediction of the Q-value which is used for policy computation, allows for a split into two networks.
				These two networks are the \emph{evaluation network},
				% $Q^{DQN}$ 
				which is used for policy computation, and the \emph{target network}, 
				% $\hat{Q}^{DQN}$ 
				which is used for predicting the target value for continuously fitting the evaluation network.
				% The evaluation network $Q^{DQN}$ also is the network that predicts the Q-value which is used by the policy $\pi$ to derive the next action choice.
				In order to keep the target network up to date, it is replaced by a copy of the evaluation network every $c$ steps.
				
				\paragraph{Double Deep Q-Network.}
			 	\noindent Deep Q-Networks tend to overestimate the prediction of Q-values for some actions,
			 	%to different degrees per action 
			 	which may result in an unjustified bias towards certain actions.
			% lead to the target network overestimating one action more than other actions and taking this one action repeatedly even though other actions might be worth to be considered.
			 	To address this problem, Double Deep Q-Networks \citeauthor{Hasselt2015} additionally use the target and evaluation networks to decouple the action choice and Q-value prediction by letting the evaluation network choose the next action to be played, and letting the target network predict the respective Q-value.
			 	% This technique of decoupling action choice and target value prediction is called \emph{Double Q-learning} and has been introduced first by \citeauthor{Hasselt2010}.

\begin{comment}

			After the utilization of these techniques, the theoretical implementation of the Deep Q-Network algorithm can be seen in algorithm \ref{dqn algorithm}. For a more detailed description of the implementation refer to \citeauthor{Mnih2013} and \cite{RL-Atari}.\\
			
			\begin{algorithm}[H]
			\KwData{MDP ($S$, $A$, $P$, $R$)}
			\KwResult{DQN and optimal policy $\pi^*$}
			$Q^{DQN}_a(s) \gets \text{arbitrary}, \forall (s, a) \in S \times A$\\
			$M \gets \text{initalize empty replay memory}$\\
			$\hat{Q}^{DQN}_a \gets Q^{DQN}_a, \forall a \in A$\\
			\For{$n \gets 1$ \KwTo \#episodes}
			{
				$s \gets s_0$\\
				\While{s not terminal}
				{
					$a \gets$ get action from policy $\pi$ in state $s$ derived from $Q^{DQN}$\\
					$r$, $s' \gets$ receive from environment by performing action $a$ in state $s$\\
					$\text{Store } (s, a, r, s') \text{ in } M$\\
					\For{$\text{random minibatch of } (s_M, a_M, r_M, s'_M) \in M$}
					{
						$a' = \argmax_a{Q^{DQN}_a(s'_M)}$\\
						$Q^{DQN}_{a_M}(s_M) = Q^{DQN}_{a_M}(s_M) + \alpha [ r_M + \gamma \hat{Q}^{DQN}_{a'}(s'_M) - Q^{DQN}_{a_M}(s_M) ]$\\
					}
					$\text{Every } c \text{ steps: } \hat{Q}^{DQN}_a \gets Q^{DQN}_a, \forall a \in A$\\
					$s \gets s'$
				}
			 }
			\caption{Deep Q-Network}
			\label{dqn algorithm}
		    \end{algorithm}
		
\end{comment}

%%%%%%%%%%%%%%%%%%%%%%%%%%%%%%%%%%%%%
	\section{Deep Ordinal % Markov Decision Process and Ordinal
	Reinforcement Learning}
	\label{ordinal definition}
	In this section, Markov decision processes and reinforcement learning algorithms are adapted to settings with ordinal reward signals. More concretely, we present
	a method for reward aggregation that fits ordinal rewards
	and explain how this method can be used in Q-learning and Deep Q-Networks in order to learn to solve environments that return feedback signals on an ordinal scale.
	
		\subsection{%Value function and policy for 
		Ordinal Markov Decision Process}
		Similar to the standard Markov Decision Process, \cite{Weng2011} defines an ordinal version of an MDP as a tuple of ($S$, $A$, $T$, $R_o$) with the only difference that $R_o$ is the reward function $S \times A \times S \rightarrow \mathbb{N}$ is modified to return ordinal rewards instead of numerical ones. Thus, it maps executing action $a$ in state $s$ and reaching state $s'$ to an ordinal reward $r_o$ from a subset of possible ordinal rewards $r_o \in \{1, ..., n\} \subset \mathbb{N}$, with $n$ representing the number of ordinal rewards.
%		
\begin{comment}

		\begin{itemize}
		\item $S$ being a finite set of states
		\item $A$ being a finite set of actions
		\item $T$ being the transition function $S \times A \times S\rightarrow \mathbb{R}$
		\item $R_o$ being the reward function $S \times A \rightarrow \mathbb{N}$ that maps executing action $a$ in state $s$ to an ordinal reward $r_o \in [1, ..., n]$, with $n$ representing the number of ordinal rewards.
		This function is different from the standard reward function since it is modified to return ordinal rewards instead of numerical ones. 
		\end{itemize}
		
\end{comment}
%
%        \hfill \break
		% When using rewards on an ordinal scale instead of numerical rewards certain properties change from one scale to the other.
		Whereas a real-valued reward provides information about the qualitative size of the reward, the ordinal scale breaks rewards down to naturally ordered \emph{reward tiers}.
		These reward tiers solely represent the rank of desirability of a reward compared to all other possible rewards, which is noted as the ranking position $r_o$ of a reward $r$ in the set of all possible rewards $\{r_1, ..., r_n\}$.
		% This way ordinal rewards are directly represented by their position between possible rewards.
		Interpreting the reward signals on an ordinal scale still allows us to order and directly compare individual reward signals, but while the numerical scale allows for comparison of rewards by means of the magnitude of their difference, ordinal rewards do not provide this information.
		
		% Since no traditional numerical operations can be done for rewards on an ordinal scale because they do not represent a magnitude but a ranking, the aggregation of ordinal rewards is different from numerical rewards.
		In order to aggregate multiple ordinal rewards, a distribution to store and represent the expected frequency of received rewards on the ordinal scale is constructed.
		This distribution is represented by a vector $D(s, a)$, in which $d_i(s, a)$ represents the frequency of receiving the ordinal reward $r_i$ by executing $a$ in $s$.
		The distribution vector is defined by
		\begin{equation} D(s, a) = \begin{bmatrix}d_1(s, a) \\ ... \\ d_n(s, a)\end{bmatrix} \end{equation}
		Through normalization of distribution vector $D$, a probability distribution $P$ can be constructed, which represents the expected probability of receiving a reward.
		The probability distribution is represented by a probability vector $P(s, a)$, in which $p_i(s, a)$ represents the estimated probability of receiving the ordinal reward $r_i$ by executing $a$ in $s$.
		Hence the probability vector can be defined by
		\begin{equation} P(s, a) = \begin{bmatrix}p_1(s, a) \\ ... \\ p_n(s, a)\end{bmatrix}
		\text{ with } \sum_{i=1}^n p_i(s, a) = 1 
		\text{ and } 0 \leq p_i(s, a) \leq 1\end{equation}
\begin{comment}

		\noindent As an example see the probability distribution for three different sets of received ordinal rewards with $P(s, a_1) = \begin{bmatrix}0.1 \\ 0.4 \\ 0.1 \\ 0.4\end{bmatrix}$, $P(s, a_2) = \begin{bmatrix}0.4 \\ 0.0 \\ 0.1 \\ 0.5\end{bmatrix}$ and $P(s, a_3) = \begin{bmatrix}0.0 \\ 0.0 \\ 1.0 \\ 0.0\end{bmatrix}$.

\end{comment}
\begin{comment}	

		\begin{figure}[htb]
		\begin{center}
		\includegraphics[scale=0.5]{Bilder/Figure_1.png}
		\end{center}
		\caption{Probability distributions of received ordinal rewards with four reward tiers}
		\label{histogram}
		\end{figure}

		For example in an environment with existing reward signals $r \in [-5.0, 0.0, 4.0, 7.5, 100.0]$, these reward signals are translated to ordinal rewards $r_o \in  [1, 2, 3, 4, 5]$ with the mapping of $r$ to $r_o$ being in the same order of ascending value $M_o: r_i \rightarrow r_{o_i}$.
		This way after the transformation of feedback signals $4.0$ and $7.5$ one still can say that $7.5 \geq 4.0$ but the information about the size of the difference $7.5-4.0$ between the individual feedback signals is discarded.
		
\end{comment}
		
		\subsubsection{Value function for ordinal rewards.}
		\label{ordinal value function}
		While numerical rewards enable the representation of value function $V_\pi(s)$ by the expected sum of rewards, the value function for environments with ordinal rewards needs to be estimated differently.
	    Since ordinal rewards are aggregated in a distribution of received ordinal rewards, the calculation of value function $V_\pi(s)$ in state $s$ can be done based on $P(s, a)$ for action $a$ that is selected by policy $\pi$.
		Hence the computation of the value function can be modeled by the following formula of
		\begin{equation}V_\pi(s) = F(P(s, a)) \text { with } a = \pi(s)
		\label{eq:V-ordinal}\end{equation}
				\noindent The computation of the value function from probability distribution $P(s, a)$ through function $F$ is performed by the technique of \emph{measure of statistical superiority}
				%, which is explained in 
				\citeauthor{Joppen2019}.
				This measure computes the probability that action $a$ receives a better ordinal reward than a random alternative action $a'$ in the same environmental state $s$.
				% An action $a$ performing better or \emph{winning} against another action $a'$ in state $s$ can be formalized as the state-action pair $(s, a)$ getting a higher reward on the ordinal scale compared to $(s, a')$.
				%The expected winning probability of action $a$ 
				This probability can be calculated through the sum of all probabilities of $a$ receiving a better ordinal reward $o$ than $a'$.
				%, for every reward tier $r_o \in [1, ..., n]$.
				Hence the probability of an action $a$ performing better than another action $a'$ can be defined as
				
				\begin{eqnarray*}\mathds{P}(a \succ a') = \sum_{o=1}^n p_o(s, a) \cdot \bigg(p_{o^<}(s, a') + \frac{1}{2} p_o(s, a')\bigg)\\
				\text{ with } p_{o^<}(s, a) = \sum_{i=1}^{o-1} p_i(s, a)\label{eq:F-ordinal}\end{eqnarray*}
				To deal with ties, additionally half the probability of $a$ receiving the same reward tier as $a'$ is added.
				
				% The winning probability of $a$ can be intuitively seen as the probability of randomly drawing a reward from both distributions $P(s, a)$ and $P(s, a')$ and getting a better ordinal reward for the former one.
				% Since naturally a higher winning probability of an action $a$ relative to other actions $a'$ is preferred, the expected winning probability $\mathds{E}[\mathds{P}(a \succ a')]$ is directly used as the value function $V(P(s, a))$.
				% In order to calculate the total expected win probability of an action $a$ against a random other action $a'$, every probability for $a$ to win against one of $k-1$ different competitor decisions $\mathds{P}(a \succ a')$ is summed up and averaged.
				The function of the measure of statistical superiority therefore is computed through the averaged winning probability of $a$ against all other actions $a'$ by
				\begin{equation}F(P(s, a)) = \mathds{E}[\mathds{P}(a \succ a')] = \frac{\sum_{a'}{\mathds{P}(a \succ a')}}{k-1}\end{equation}
				for $k$ available actions in state $s$.
		Based on \eqref{eq:V-ordinal}, the optimal policy $\pi^*$ can be determined in the same way as for numerical rewards~\eqref{eq:V-pi-star} by maximizing the respective value function $V_\pi(s)$.

	    \subsection{Transformation of existing numerical rewards to ordinal rewards}
		If an environment has pre-defined rewards on a numerical scale, transforming numerical rewards $r \in \{r_1, ..., r_n\} \subset \mathbb{R}$ into ordinal rewards $r_o \in \{1, ..., n\} \subset \mathbb{N}$ can easily be done by translating every numerical reward to its ordinal position within all possible numerical rewards.
		This way the lowest possible numerical reward is mapped to position 1, and the highest numerical reward is mapped to position $n$, with $n$ representing the number of possible numerical rewards.
		This transformation process simply results in removing the metric and semantic of distances of rewards but keeping the order.
		% This can be modelled by a mapping $M_o: \mathbb{R} \rightarrow \mathbb{N}$ which transforms the numerical rewards to their ranking positions in ascending order.

		\subsection{Ordinal Reinforcement Learning}
		\label{ordinal reinforcement learning}
		In Section~\ref{ordinal value function}, we have shown how to compute a value function $V_\pi(s)$ and defined the optimal policy $\pi^*$ for environments with ordinal rewards.
		This can now be used for adapting common reinforcement learning algorithms to ordinal rewards.
		% We adapt all reinforcement algorithms presented in \ref{reinforcement learning} to \emph{ordinal reinforcement learning}.
					
			\subsubsection{Ordinal Q-learning.}
			\label{ordinal qlearning}
			For the adaptation of the Q-learning algorithm to ordinal rewards, we do not directly update a Q-value $Q(s, a)$ that represents the quality of a state-action pair $(s, a)$ but update the distribution $D(s, a)$ of received ordinal rewards.
			The target distribution is computed by adding the received ordinal reward $i$ (represented through unit vector $e_i$ of length $n$) to the distribution $D(s', \pi^*(s'))$ of taking an action in the new state $s'$ according to the optimal policy $\pi^*$.
			The previous distribution $D(s, a)$ is updated with the target distribution by interpolating both values with learning rate $\alpha$, which can be seen in the formula
			\begin{equation}D(s, a) = D(s, a) + \alpha [ e_i(s, a) + \gamma D(s', \pi^*(s')) - D(s, a) ]\label{eq:ordinal-q-update}\end{equation}
%			
			% $P(s, a)$ represents the probability distribution of receiving future ordinal rewards when taking action $a$ in state $s$ and following the optimal policy $\pi^*$ afterwards.
			In this adaptation of Q-learning\footnote{This technique of modifying the Q-learning algorithm to deal with rewards on an ordinal scale can analogously be applied to other Q-table based reinforcement learning algorithms like \emph{Sarsa} and \emph{Sarsa-$\lambda$} \cite{Zap2019}}, the expected quality of state-action pair $(s, a)$ is not represented by the Q-value $Q(s, a)$ \eqref{eq:q-update} but by the function $F(P(s, a))$ \eqref{eq:F-ordinal} of the probability distribution $P(s, a)$, which is derived from the iteratively updated distribution $D(s, a)$.

			\subsubsection{Ordinal Deep Q-Network.}		
			\label{ordinal dqn}	
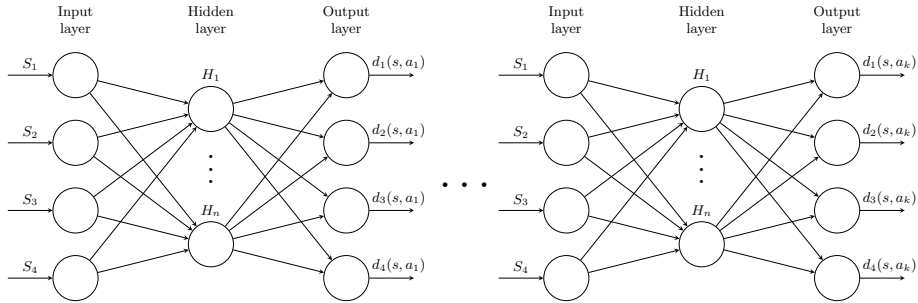
\begin{figure}[t]
			\scalebox{.6}{
			\begin{tikzpicture}[x=1.5cm, y=1.5cm, >=stealth]

			\foreach \m/\l [count=\y] in {1,2,3,4}
			  \node [every neuron/.try, neuron \m/.try] (input-\m) at (0,2.5-\y) {};
			
			\foreach \m [count=\y] in {1,missing,2}
			  \node [every neuron/.try, neuron \m/.try ] (hidden-\m) at (2,2-\y) {};
			
			\foreach \m [count=\y] in {1,2,3,4}
			  \node [every neuron/.try, neuron \m/.try ] (output-\m) at (4,2.5-\y) {};
			
			\foreach \l [count=\i] in {1,2,3,4}
			  \draw [<-] (input-\i) -- ++(-1,0)
			    node [above, midway] {$S_\l$};
			
			\foreach \l [count=\i] in {1,n}
			  \node [above] at (hidden-\i.north) {$H_\l$};
			
			\foreach \l [count=\i] in {1,2,3,4}
			  \draw [->] (output-\i) -- ++(1,0)
			    node [above, midway] {$ \quad d_{\l}(s, a_1)$};
			
			\foreach \i in {1,...,4}
			  \foreach \j in {1,...,2}
			    \draw [->] (input-\i) -- (hidden-\j);
			
			\foreach \i in {1,...,2}
			  \foreach \j in {1,...,4}
			    \draw [->] (hidden-\i) -- (output-\j);
			
			\foreach \l [count=\x from 0] in {Input, Hidden, Output}
			  \node [align=center, above] at (\x*2,2) {\l \\ layer};
			  
			%%%
			
			\foreach \m [count=\y] in {hmissing}
			  \node [every neuron/.try, neuron \m/.try ] (hidden-\m) at (5.75,0) {};
			
			%%%
			  
			\foreach \m/\l [count=\y] in {1,2,3,4}
			  \node [every neuron/.try, neuron \m/.try] (input-\m) at (7.25,2.5-\y) {};
			
			\foreach \m [count=\y] in {1,missing,2}
			  \node [every neuron/.try, neuron \m/.try ] (hidden-\m) at (9.25,2-\y) {};
			
			\foreach \m [count=\y] in {1,2,3,4}
			  \node [every neuron/.try, neuron \m/.try ] (output-\m) at (11.25,2.5-\y) {};
			
			\foreach \l [count=\i] in {1,2,3,4}
			  \draw [<-] (input-\i) -- ++(-1,0)
			    node [above, midway] {$S_\l$};
			
			\foreach \l [count=\i] in {1,n}
			  \node [above] at (hidden-\i.north) {$H_\l$};
			
			\foreach \l [count=\i] in {1,2,3,4}
			  \draw [->] (output-\i) -- ++(1,0)
			    node [above, midway] {$ \quad d_{\l}(s, a_k)$};
			
			\foreach \i in {1,...,4}
			  \foreach \j in {1,...,2}
			    \draw [->] (input-\i) -- (hidden-\j);
			
			\foreach \i in {1,...,2}
			  \foreach \j in {1,...,4}
			    \draw [->] (hidden-\i) -- (output-\j);
			
			\foreach \l [count=\x from 0] in {Input, Hidden, Output}
			  \node [align=center, above] at (\x*2+7.25,2) {\l \\ layer};

			\end{tikzpicture}}
			\caption{Example of an array of ordinal deep neural networks for DQN for reward distribution prediction}
			\label{ordinal neural net}
			\end{figure}

			Because ordinal rewards are aggregated by a distribution instead of a numerical value, the neural network is adapted to predict distributions $D(s, a)$ instead of Q-values for every possible action.
			Hence for one action the network does not predict a 1-dimensional Q-value, but predicts an $n$-dimensional reward distribution with $n$ being the length of the ordinal scale.
			Since this distribution has to be computed for each of $k$ actions, the adaptation of the Deep Q-Network algorithm to ordinal rewards requires a differently structured neural network.
            Contrary to the original Deep Q-Network where one network simultaneously predicts $k$ Q-values for all actions, the structure of the ordinal DQN consists of an array of $k$ neural networks, from which every network computes the expected ordinal reward distribution $D(s, a)$ for one separate action $a$. 
			In a deep neural network for the prediction of distributions every output node of the network computes one distribution value $d_i(s, a)$.
			% Since probability values of every ordinal reward tier have to be computed for one action, the output of each neural network is $n$-dimensional, with every node representing the reward probability $p_i$ of one of $n$ ordinal reward tiers.
			The structure of neural networks used for the prediction of distributions can be seen in Figure~\ref{ordinal neural net}.

%			\hfill \break
			The prediction of the ordinal reward distributions $D(s, a)$ for all actions can afterwards be normalized to a probability distribution and used in order to compute the value function $V_\pi(s)$ through the measure of statistical superiority as has been previously defined in \eqref{eq:V-ordinal}. %\ref{ordinal value function}.
			Once the value function and policy have been evaluated, the ordinal variant of the DQN algorithm follows a similar procedure as ordinal Q-learning and updates the prediction of the reward distribution for $(s, a)$ by fitting $D^{DQN}_{a}(s)$ to the target reward distribution:
			\begin{equation}\hat{D}^{DQN}_{a}(s) = e_{r_o}(s, a) + \gamma D^{DQN}_{\pi^*(s')}(s')
			\end{equation}
			
			The main difference in the update step between ordinal Q-learning \eqref{eq:ordinal-q-update} and ordinal DQN consists of fitting the neural network of action $a$ for input $s$ to the expected reward distribution by backpropagation instead of updating a Q-table entry $(s, a)$.
			Additional modifications to the ordinal Deep Q-Network in form of experience replay, the split of the target and evaluation network and the usage of a Double DQN are done in a similar fashion as described with the standard DQN algorithm in Section \ref{dqn}.
			These modifications can be seen in the following paragraphs.
			
			\paragraph{Experience replay.}
			% \noindent The concept of \emph{experience replay} is done in the same manner for the ordinal variant as in the standard DQN algorithm.
			% \noindent A memory $M$ of previous experiences is saved by storing a tuple of state $s$, chosen action $a$, received ordinal reward $r_o$ and the newly reached state $s'$ for every executed decision.
			\noindent A memory $M$ is used to sample multiple saved experience elements $(s, a, r_o, s')$ randomly and replay these previously seen experiences by fitting the ordinal DQN networks to the samples of earlier memory elements.
			% This way the overfitting of the network to recent states and actions, which are highly correlated across multiple time steps, is prevented.
			
			\paragraph{Target and evaluation networks.}
			\noindent In order to prevent unstable behavior by using the same networks for the prediction and updating step, we use separate evaluation networks to predict reward distributions for the policy computation, and use target networks to predict the target reward distributions which are used for fitting the evaluation networks continuously.
			% The computation of the probability distributions $P$ for the value function $V(P(s, a))$ for action choice by policy $\pi$ is done by the continuously fitted evaluation networks $P^{DQN}$.
			% In order to periodically update the target networks, the weights of every evaluation network are copied to the respective target network every $c$ steps.
			
			\paragraph{Double Deep Q-Network.}
			\noindent The neural networks of ordinal DQN tend to overestimate the prediction of the reward distributions for some actions, which may result in an unjustified bias towards certain actions.
			% This issue is solved in standard DQN by decoupling the action choice and target value prediction in the updating step to be made separately by target and evaluation networks.
			% In order to use the technique of Double Deep Q-Network for the ordinal DQN, the decoupling of these two tasks has to be done similarly by determining the action choice based on the evaluation network and predicting the target value by the target network.
		    Therefore, in order to determine the next action to be played by $\pi^*$, the measure of statistical superiority is computed based on the reward distributions predicted by the evaluation networks.
			Afterwards the prediction of the reward distribution for this action is computed by the respective target network.
%
\begin{comment}
			
			\hfill \break
			\noindent The pseudo-code of the final \emph{ordinal Deep Q-Network} with all above modifications can be seen in algorithm \ref{ordinal dqn algorithm}.\\
			
			\begin{algorithm}[H]
			\KwData{MDP ($S$, $A$, $P$, $R_o$)}
			\KwResult{DQN and optimal policy $\pi^*$}
			$P^{DQN}_a(s) \gets \text{arbitrary}, \forall (s, a) \in S \times A$\\
			$M \gets \text{initalize empty replay memory}$\\
			\For{$n \gets 1$ \KwTo \#episodes}
			{
				$s \gets s_0$\\
				\While{s not terminal}
				{
					$a \gets$ get action from policy $\pi$ in state $s$ derived from $P^{DQN}$\\
					$r_o$, $s' \gets$ receive from environment by performing action $a$ in state $s$\\
					$\text{Store } (s, a, r_o, s') \text{ in } M$\\
					\For{$\text{random minibatch of } (s_M, a_M, r_M, s'_M) \in M$}
					{
						$a' \gets$ get action from optimal policy $\pi^*$ in state $s'_M$ derived from $P^{DQN}$\\
						$P^{DQN}_{a_M}(s_M) = P^{DQN}_{a_M}(s_M) + \alpha [ e_{r_M} + \gamma \hat{P}^{DQN}_{a'}(s'_M) - P^{DQN}_{a_M}(s_M) ]$\\
					}
					$\text{Every } c \text{ steps: } \hat{Q} \gets Q$\\
					$s \gets s'$
				}
			 }
			\caption{Ordinal Deep Q-Network}
			\label{ordinal dqn algorithm}
		    \end{algorithm}
		    
\end{comment}
%
%%%%%%%%%%%%%%%%%%%%%%%%%%%%%%%%%%%%%
	\section{Experiments and Results}
	\label{experiments}
	In the following, the standard reinforcement algorithms described in Section \ref{reinforcement learning} and the ordinal reinforcement learning algorithms described in Section \ref{ordinal reinforcement learning} are evaluated and compared in a number of testing environments.\footnote{The source code for the implementation of the experiments can be found in \url{https://github.com/az79nefy/OrdinalRL}.}
	% At first the detailed experimental setup is explained and afterwards the results from said setup are presented and investigated.
		\subsection{Experimental setup}
		The environments which are used for evaluation are provided by OpenAI Gym,\footnote{For further information about OpenAI visit  \url{https://gym.openai.com}.} which can be viewed as a unified toolbox for our experiments.
		% OpenAI Gym provides a number of environments that can be used for testing and evaluating reinforcement learning algorithms. 
		All environments expect an action input after every time step and return feedback in form of the newly reached environmental state, the direct reward for the executed action, and the information whether the newly reached state is terminal.
		%Furthermore all environments are deterministic, meaning that all transition functions $T$ of the Markov Decision Processes in the environment are non-probabilistic with every transition having a probability of 1.
		%Additionally the environments are static, hence they do not change internally over time, with the reward and transition function $R$ and $T$ remaining static over the course of all time steps. 
		%Therefore the algorithms should converge to an optimal behavior if the reinforcement learning algorithm is successful.
		The environments that the algorithms were tested on were \emph{CartPole} and \emph{Acrobot}.\footnote{Further technical details about the environments CartPole and Acrobot from OpenAI can be found in \url{https://gym.openai.com/envs/CartPole-v0/} and \url{https://gym.openai.com/envs/Acrobot-v1/}.}
		
		Policies of the reinforcement learning algorithms were modified to use \mbox{$\epsilon$-greedy} exploration \citeauthor{Sutton2018}, which encourages early exploration of the state space and increases exploitation of the learned policy over time. 
		In the experiments the maximum exploitation is reached after half of the total episodes.
		In order to directly compare the standard and the ordinal variants of reinforcement learning algorithms, the quality of the learned policy and the computational efficiency are investigated across all environments with varying episode numbers.
		Information about the quality of the learned policy is derived from the sum of rewards over a whole episode (score) or the win rate while the efficiency is measured by real-time processing time.
		Additionally to the standard variant with unchanged rewards, the performance of standard Q-learning algorithms is tested with changed rewards in order to simulate the performance on environments where no optimal reward engineering has been performed.
		It should be noted that the modifications of the rewards is performed under the constraints of remaining existing reward order, therefore not changing the transformation to the ordinal scale.
		The change of rewards (CR) from the existing numerical rewards $r \in \{r_1, ..., r_n\}$ is performed for all rewards by the calculation of $r_{_{CR, i}} = \frac{r_i - min(r)}{100}$.

		The parameter configuration of the Q-learning algorithms is learning rate $\alpha=0.1$ and discount factor $\gamma=0.9$.
		The parameter configuration of the Deep Q-Network algorithm is learning rate $\alpha=0.0005$ and discount factor $\gamma=0.9$. As for the network specific parameters, the \emph{Adam} optimizer is used for the network fitting, the target network is getting replaced every 300 fitting updates, the experience memory size is 200000 and the replay batch size is 64.

		\subsection{Experimental results}
		The results of the comparison between numerical and ordinal algorithms for the CartPole- and Acrobot-environment in terms of score, win rate and computational time are shown and investigated in the following. This comparison is performed based on the averaged results from 10 and respectively 5 independent runs of Q-learning and Deep Q-Network on the environments.
		
		\subsubsection{Q-learning.}
%			
\begin{comment}

			\begin{figure}[!ht]
			   \begin{minipage}{\textwidth}
			     \centering
			     \includegraphics[width=.4\textwidth]{Bilder/Evaluation/cartpole/q_learning/cartpole_qlearning_400_rate.png}\quad
			     \includegraphics[width=.4\textwidth]{Bilder/Evaluation/cartpole/q_learning/cartpole_qlearning_2000_rate.png}\\
			     \includegraphics[width=.4\textwidth]{Bilder/Evaluation/cartpole/q_learning/cartpole_qlearning_10000_rate.png}\quad
			     \includegraphics[width=.4\textwidth]{Bilder/Evaluation/cartpole/q_learning/cartpole_qlearning_50000_rate.png}
			     \caption{Win rate comparison of standard and ordinal Q-learning for varying episode numbers}
			     \label{fig:sub1}
			   \end{minipage}\\[1em]
			\end{figure}
			
\end{comment}
%

			\begin{figure}[t]
			   \begin{minipage}{\textwidth}
			     \centering
			     \includegraphics[width=.39\textwidth]{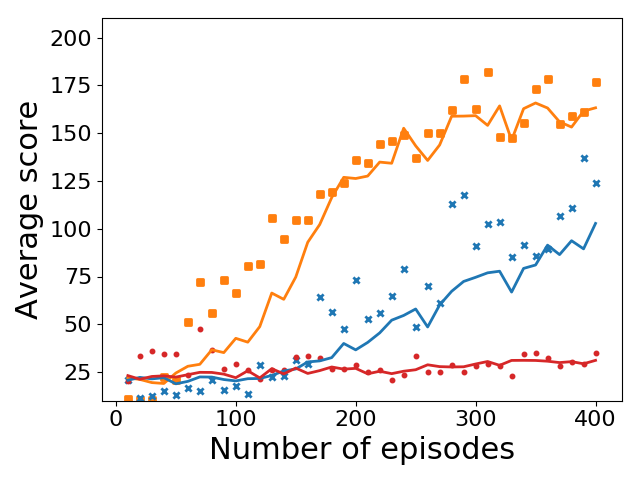}
			     \includegraphics[width=.39\textwidth]{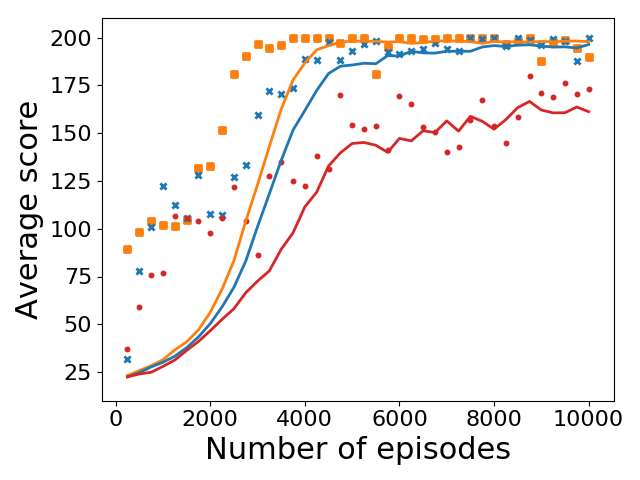}
			     \raisebox{0.8\height}{\includegraphics[width=.20\textwidth]{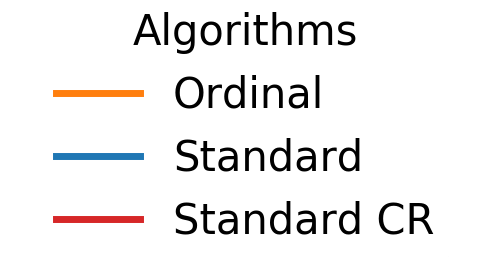}}
			     \caption{CartPole scores of standard and ordinal Q-learning for 400 and 10000 episodes}
			     \label{fig:sub21}
			   \end{minipage}
			\end{figure}
			
			In Figure \ref{fig:sub21} the scores for the CartPole-environment over the course of 400 and 10000 episodes can be seen which were played by an agent using the ordinal (orange) as well as the standard Q-learning algorithm, with (red) and without (blue) modified rewards. 
			Additionally the individual dots in this figure represent the scores achieved by the respective algorithms by using the optimal policy instead of $\epsilon$-greedy exploration.
			The evaluation of these scores shows that the ordinal variant of Q-learning performs better than the standard variant with engineered rewards for 400 episodes and reaches the optimal score of 200 quicker for 10000 episodes.
			Additionally the use of ordinal rewards significantly outperforms the standard variant with modified rewards for both episode numbers.
			Therefore it can be seen that ordinal Q-learning is able to learn a good policy better than the standard variants for the CartPole-environment.
			% The same observation can be made in figure \ref{fig:sub1} for the win rates.
			
			\begin{figure}[t]
			   \begin{minipage}{\textwidth}
			     \centering
			     \includegraphics[width=.4\textwidth]{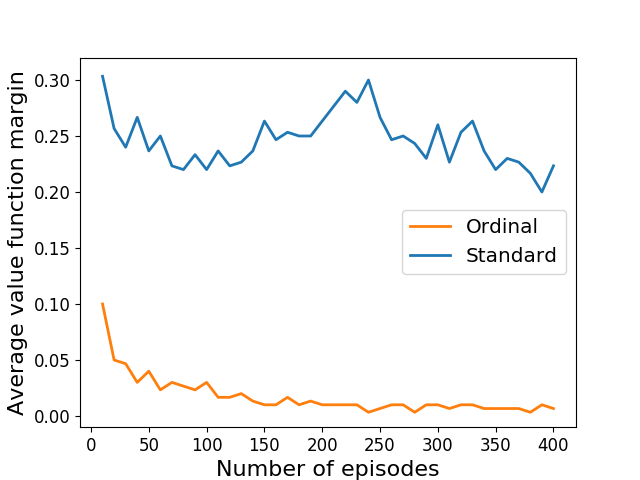}\quad
			     \includegraphics[width=.4\textwidth]{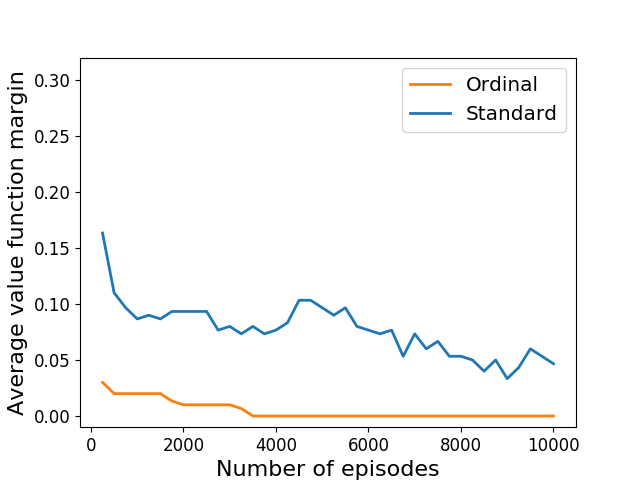}
			     \caption{Comparison of value function margin for best action of standard and ordinal Q-learning for 400 and 10000 episodes of CartPole}
			     \label{fig:sub6}
			   \end{minipage}\\[1em]
			\end{figure}
			
			In order to explain the difference of learned behavior between the standard and ordinal variant, the average relative difference of Q-values $Q(s, a)$ and respectively measure of statistical superiority functions $F(P(s, a))$ for the two possible actions were plotted and compared in Figure \ref{fig:sub6} for standard (blue) and ordinal (orange) Q-learning.
			It can be seen for both episode numbers that the policy which is learned by ordinal RL through the measure of statistical superiority converges to a difference of 0, meaning that the function $F(P(s, a))$ converges to similar values for both actions.
			This can be interpreted as the policy learning to play safely and rarely entering any critical states where this function would indicate strong preference towards one action (e.g. in big angles).
			On the other side it can be seen for 400 episodes that common RL does not converge towards similar Q-values for the actions over time and therefore a policy is learned that enters critical states more often.
			It should be noted that the Q-value differences for standard Q-learning converges to 0 for evaluations with more episodes and a safe policy is eventually learned as well.
			
			\begin{figure}[t]
			   \begin{minipage}{\textwidth}
			     \centering
			     \includegraphics[width=.39\textwidth]{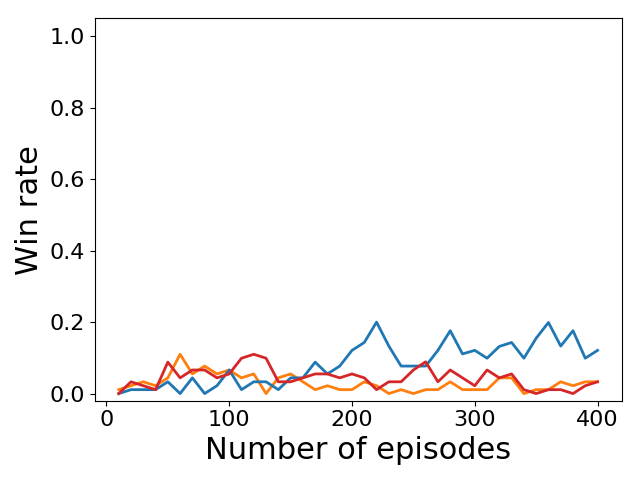}
			     \includegraphics[width=.39\textwidth]{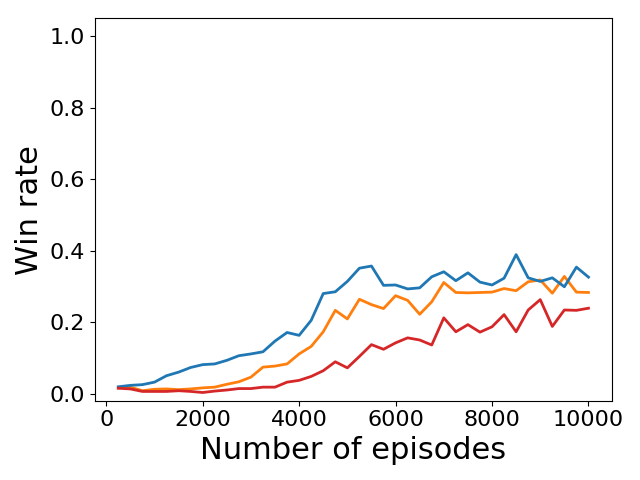}
			     \raisebox{0.8\height}{\includegraphics[width=.20\textwidth]{Bilder/Evaluation/legend_algorithms.png}}
			     \caption{Acrobot win rates of standard and ordinal Q-learning for 400 and 10000 episodes}
			     \label{fig:sub22}
			   \end{minipage}
			\end{figure}
			
	       In Figure \ref{fig:sub22} the win rates from the Acrobot-environment were plotted over the course of 400 and 10000 episodes similarly as the scores for the CartPole-environment and it can be seen for low episode numbers that while the policy learned by the standard variant of Q-learning with unchanged rewards performs better than the policy learned by the ordinal variant, changing the numerical values of rewards yields the same performance as the ordinal variant.
	       But for high episode numbers it should be noted that the ordinal variant reaches a similar performance as the standard variant with a win rate of 0.3 after 10000 episodes and clearly outperforms the win rate of the standard Q-learning algorithm with CR.     
	       	\begin{figure}[t]
			   \begin{minipage}{\textwidth}
			     \centering
			     \includegraphics[width=.4\textwidth]{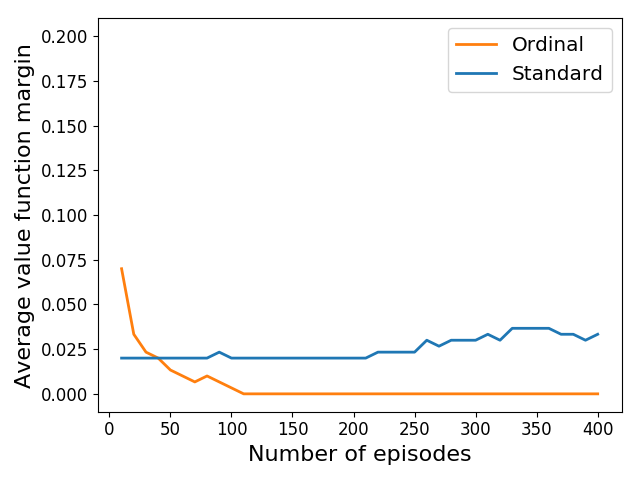}\quad
			     \includegraphics[width=.4\textwidth]{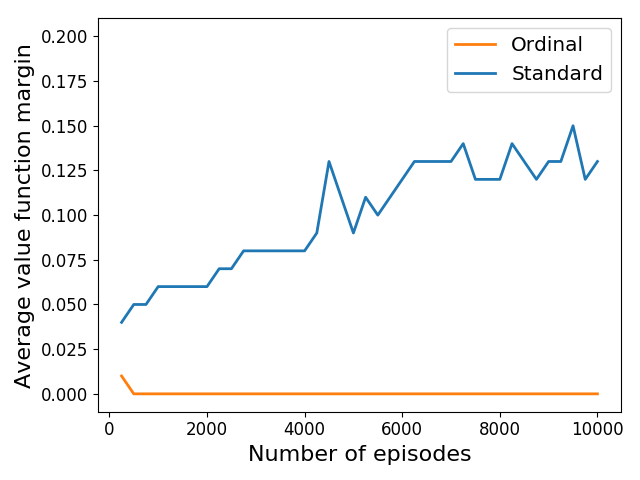}
			     \caption{Comparison of value function margin for best action of standard and ordinal Q-learning for 400 and 10000 episodes of Acrobot}
			     \label{fig:sub7}
			   \end{minipage}\\[1em]
			\end{figure}
	       
	       Similar as for the CartPole-environment, the $F$- and $Q$-function margins of the best actions over the course of 400 and 10000 episodes were compared in Figure \ref{fig:sub7} and yield different observations for the standard and ordinal variants, and it can be therefore be concluded that the learned policies differ.
	       While the ordinal variant decreases the relative margin of $F(P(s, a))$ of the best action and therefore learns a policy which plays safely, the standard variant learns a policy which maximizes the Q-value margin of the best action and therefore follows a policy which enters critical states more often.
	       While the standard variant learns a good policy quicker, it should be noted that both policies perform comparably after many episodes despite the policy differences.
	       
	       		\begin{table}[t]
						    \caption{Computation time comparison of standard and ordinal Q-learning for varying episode numbers}
			    \label{times}
			   
			   \centering \begin{tabular}
%			    { | p{1.5cm} || p{2.5cm} | p{2.5cm} || p{2.5cm} | p{2.5cm} | }
			    {l@{\hskip 12pt}r@{\hskip 12pt}r@{\hskip 12pt}r@{\hskip 12pt}r}
			    \toprule
			    Number of &  \multicolumn{2}{c}{CartPole} &  \multicolumn{2}{c}{Acrobot} \\ % \hline
			   episodes  & Standard % variant 
			   & Ordinal 
			   % variant 
			   & Standard % variant 
			   & Ordinal %variant 
			   \\ \midrule
			    400 & 2.10 s & 4.17 s & 35.74 s & 52.85 s \\ % \hline
			    2000 & 10.07 s & 24.86 s & 174.38 s & 266.40 s \\ %\hline
			    10000 & 67.29 s & 130.09 s & 855.15 s & 1258.30 s \\ %\hline
			    50000 & 354.52 s & 667.87 s & 4149.78 s & 6178.76 s \\ % \hline
			    \bottomrule
			    \end{tabular}
			\end{table}
	       
	       As can be seen in Table \ref{times}, using the ordinal variant results in an additional computational load by a factor between 0.8 and 1.2 for CartPole and 0.5 for Acrobot.
	       The additionally required computational capacity is caused by the computation of the measure of statistical superiority which is less efficient than computing the expected sum of rewards.
	       This factor could be reduced by using the iterative update of the function \emph{measure of statistical superiority} described in \cite{Joppen2019}.

		\subsubsection{Deep Q-Network.}
		
\begin{figure}[t]
		   \begin{minipage}{\textwidth}
		     \centering
		     \includegraphics[width=.39\textwidth]{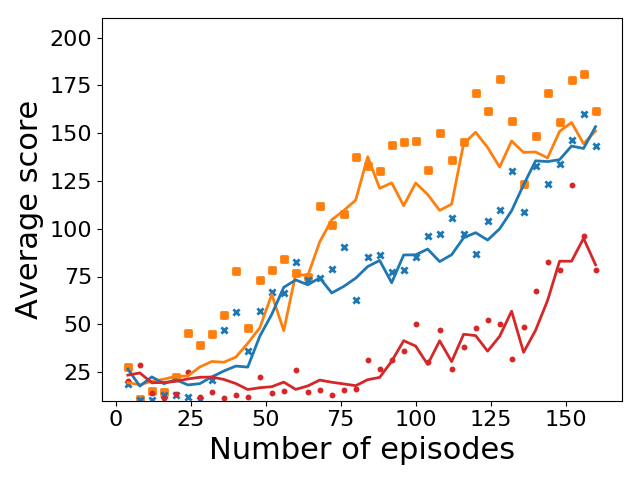}
		     \includegraphics[width=.39\textwidth]{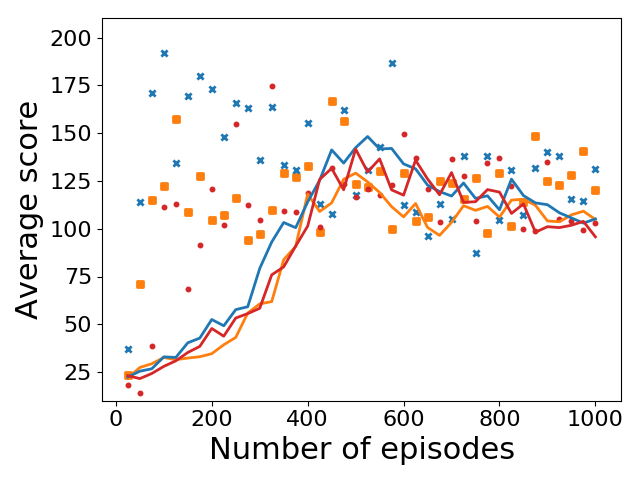} \raisebox{0.8\height}{\includegraphics[width=.20\textwidth]{Bilder/Evaluation/legend_algorithms.png}}
		     \caption{CartPole scores of standard and ordinal DQN for 160 and 1000 episodes}
		     \label{fig:sub8}
		   \end{minipage}
		\end{figure}

		In Figure \ref{fig:sub8} the scores achieved in the CartPole-environment by the ordinal as well as the standard Deep Q-Network, with and without CR, can be seen over the course of 160 and 1000 episodes.
		% The scores achieved by the optimal policy are plotted by individual dots and scores of $\epsilon$-greedy exploration are plotted by lines.
		For 160 episodes it can be seen that ordinal DQN as well as the standard variant without CR converge to a good policy reaching an episode score close to 150. 
		Contrary to this performance, modified rewards negatively impact standard Q-learning and therefore its performance is significantly worse, not reaching a score above 100.
		Additionally for low episode numbers it should be noted that the policy learned by the ordinal variant of Deep Q-Network is able to achieve good scores faster than the standard variant, matching the observation made for the Q-learning algorithms.
		The evaluation for 1000 episodes shows that the performances of standard, with and without CR, and ordinal DQNs are comparable.
		
\begin{figure}[t]
		   \begin{minipage}{\textwidth}
		     \centering
		     \includegraphics[width=.39\textwidth]{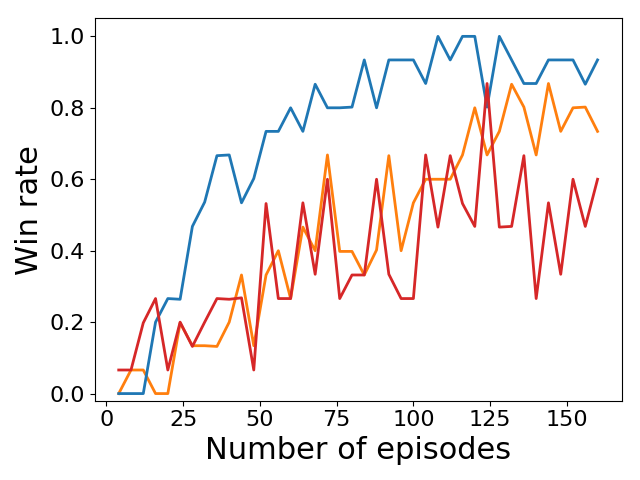}
		     \includegraphics[width=.39\textwidth]{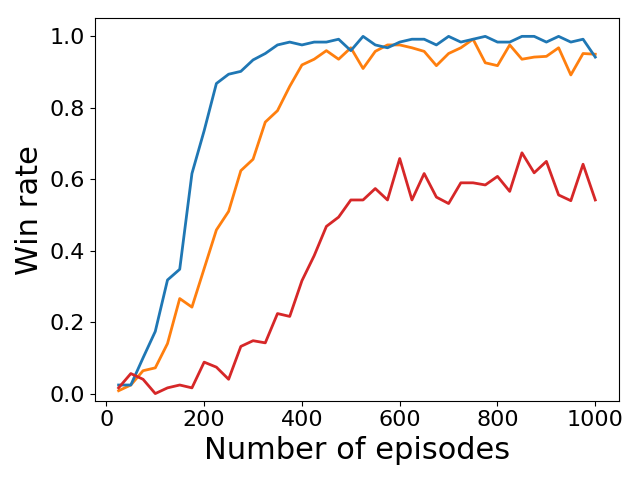}
		     \raisebox{0.8\height}{\includegraphics[width=.20\textwidth]{Bilder/Evaluation/legend_algorithms.png}}
		     \caption{Acrobot win rates of standard and ordinal DQN for 160 and 1000 episodes}
		     \label{fig:sub9}
		   \end{minipage}
		\end{figure}

		Figure \ref{fig:sub9} plots the win rate of Deep Q-Network algorithms for the Acrobot-environment over the course of 160 and 1000 episodes.
		For 160 episodes standard DQN with engineered rewards performs better than the ordinal variant, but loses this quality once the rewards are modified.
		For high episode numbers it can be seen that the ordinal variant is comparable to the standard algorithm without CR and solves the environment with a win rate of close to 1.0, but clearly outperforms the standard DQN with modified rewards which is only able to achieve a win rate of 0.6.
		It should be noted that all variants of DQN are able to learn a better policy than their respective Q-learning algorithms, achieving a higher win rate in less than 160 episodes.
		
		Additionally, it should be noted that the use of the ordinal variant of DQN adds an additional computational factor between 0 and 0.5 for the CartPole-environment and 1.0 for the Acrobot-environment, as can be seen in Table \ref{dqn times}.
		
		Since the evaluation of the ordinal Deep Q-Network algorithm shows comparable results to the standard DQN with engineered rewards and furthermore outperforms the standard variant with modified rewards, it can be concluded that the conversion of the Deep Q-Network algorithm to ordinal rewards is successful.
		Therefore it has been shown that algorithms of deep reinforcement learning can as well be adapted to the use of ordinal rewards.
		
		% It should be noted that common and deep ordinal reinforcement learning algorithms reaching a performance comparable to the use of numerical rewards is a remarkable result, since the information used by ordinal algorithms is reduced by discarding the quantitative measure of numerical rewards.
		% Additionally it has been shown in the experiments that ordinal reinforcement learning variants are able to improve the performance of the learned policy compared to standard RL algorithms for environments without engineered rewards.
	       
		\begin{table}[t]
		\caption{Computation time comparison of standard and ordinal DQN %for varying episode numbers
		}
		\label{dqn times}
		    \centering \begin{tabular}			    
		    {l@{\hskip 12pt}r@{\hskip 12pt}r@{\hskip 12pt}r@{\hskip 12pt}r}
			\toprule
			Number of &  \multicolumn{2}{c}{CartPole} &  \multicolumn{2}{c}{Acrobot} \\ % \hline
			episodes  & Standard % variant 
			& Ordinal 
			% variant 
			& Standard % variant 
			& Ordinal %variant 
			\\ \midrule
		    160 & 1520.01 s & 2232.48 s & 3659.44 s & 7442.49 s \\
		    400 & 6699.69 s & 7001.79 s & 9678.80 s & 19840.88 s \\
		    1000 & 15428.41 s & 15526.84 s & 23310.36 s & 47755.90 s \\
		    \bottomrule
		    \end{tabular}
		\end{table}

%%%%%%%%%%%%%%%%%%%%%%%%%%%%%%%%%%%%%	
\begin{comment}

	\section{Future work}
	\label{future work}
	- Use Reference-points and Quantile Criterion for comparison
	
	- Use different policies for exploration vs. exploitation trade-off
	
	- Use different method to update estimatation of ordinal reward probability distribution (take into account uncertainty of estimated distribution)
	            
\end{comment}

%%%%%%%%%%%%%%%%%%%%%%%%%%%%%%%%%%%%%	
	\section{Conclusion}
	In this paper we have shown that the use of ordinal rewards for reinforcement learning is able to reach and even improve the quality of standard reinforcement learning algorithms with numerical rewards.
	We compared RL algorithms for both numerical and ordinal rewards on a number of tested environments and demonstrated that the performance of the ordinal variant is mostly comparable to the learned common RL algorithms that make use of engineered rewards while being able to significantly improve the performance for modified rewards.
	% This is achieved by the function of measure of statistical superiority which has been specifically designed and optimized for ordinal rewards.
	% In the cases where reinforcement learning with ordinal rewards is not able to improve upon the algorithms for numerical rewards, it should be noted that the ordinal variant still converges towards the good policy with more episodes.
	% The evaluation of policy quality through the direct comparison of RL algorithms with ordinal rewards to standard variants is most notable, since quantitative scores of numerical rewards are reduced to the ordinal scale for ordinal reinforcement learning algorithms and therefore the observed comparable quality between common and ordinal RL is a remarkable result.
	
	% Anyway it should be noted that reinforcement learning based on ordinal rewards can not have same functionality as standard reinforcement learning since ordinal rewards do not capture a quantitative measure of rewards and therefore are not able to learn a desired behavior as fast as with numerical rewards.
	% Policies that rely on complete prevention of a certain environmental state or other predefined desires mostly cannot be captured directly by the ordinal scale, because the modelling of such environmental states requires the integration of rewards with high quantity or infinite rewards.
	
    Finally, it should be noted that ordinal reinforcement learning enables the learning of a 
    %good policy for any desired environment without much effort to manually shape rewards.
    good policy for environments without much effort to manually shape rewards.
    We hereby lose the possibility of reward shaping to the same degree that numerical rewards would allow, but therefore gain a more simple-to-design reward structure.
    Hence, our variant of reinforcement learning with ordinal rewards is especially suitable for environments that do not have a natural semantic of numerical rewards or where reward shaping is difficult.
    Additionally this method enables the usage of new and unexplored environments for RL only with the specification of an order of desirability instead of the needed effort of manually engineering numerical rewards with sensible semantic meaning.

%%%%%%%%%%%%%%%%%%%%%%%%%%%%%%%%%%%%%

    \section*{Acknowledgements}
    This work was supported by DFG.
	Calculations for this research were conducted on the Lichtenberg high performance computer of the TU Darmstadt.
		
\bibliographystyle{splncs04}
\bibliography{ordinalBandits}

\end{document}